\pdfoutput=1
\def\PaperShowRevisions{0}
\def\PaperBuildSupplement{2}

\RequirePackage{fix-cm}
\documentclass[10pt,twocolumn,letterpaper]{article}

\PassOptionsToPackage{table}{xcolor}
\ifnum\PaperBuildSupplement=2\relax
  \usepackage[pagenumbers]{styles/cvpr}
\else
  \usepackage[review]{styles/cvpr}
\fi
\usepackage{latexml}

\usepackage{amsfonts,amsthm,mathtools}
\usepackage[utf8]{inputenc}
\usepackage[T1]{fontenc}
\usepackage{nicefrac}
\usepackage{microtype}
\usepackage{multirow}
\usepackage{algorithm}
\usepackage{algorithmic}
\usepackage{needspace}
\usepackage{flafter}
\usepackage{stfloats}
\usepackage{placeins}
\usepackage[bottom]{footmisc} %
\usepackage{tabularx}
\newcolumntype{Y}{>{\raggedright\arraybackslash}X}

\iflatexml
  \newcounter{promptbox}
  
\else
  \usepackage[most]{tcolorbox}
  \definecolor{promptframe}{HTML}{6B6B6B}
\definecolor{promptback}{HTML}{F7F8FB}
\newtcolorbox[auto counter]{promptbox}[2][]{%
  enhanced, breakable,
  colframe=promptframe, colback=promptback,
  colbacktitle=promptframe, coltitle=white,
  boxrule=0.8pt, arc=2.5pt,
  left=5pt, right=5pt, top=4pt, bottom=4pt,
  toptitle=2pt, bottomtitle=2pt,
  fonttitle=\normalfont\small,
  fontupper=\footnotesize,
  title={Prompt~\thetcbcounter: #2},
  before upper={\setlength{\parindent}{0pt}\setlength{\parskip}{1pt}},
  before skip=6pt, after skip=8pt,
  #1}

\fi
\usepackage{enumitem}

\definecolor{cvprblue}{rgb}{0.21,0.49,0.74}
\definecolor{i2vtrow}{HTML}{F9FFF9}
\newcommand{\PaperInput}[1]{%
  \def\PaperInputPath{#1}%
  \ifnum\PaperShowRevisions=1\relax\IfFileExists{revisions/marked/#1.tex}{\def\PaperInputPath{revisions/marked/#1}}{}\fi
  \input{\PaperInputPath}}
\usepackage{trimclip}
\newsavebox{\clipwidthbox}
\newenvironment{clipwidth}{\begin{lrbox}{\clipwidthbox}}{\end{lrbox}\clipbox{0pt 0pt 0pt 0pt}{\usebox{\clipwidthbox}}}
\definecolor{i2vtgain}{rgb}{0.05,0.42,0.05}
\definecolor{i2vtdrop}{RGB}{160,0,0}

\newlength{\scorewidth}
\newlength{\deltawidth}
\def\deltaabs#1#2\relax{#2}
\newcommand{\printdelta}[1]{(%
  \ifdim #1pt>0pt\makebox[0.6em][c]{+}\makebox[1.75em][r]{\deltaabs#1\relax}%
  \else\ifdim #1pt<0pt\makebox[0.6em][c]{\textminus}\makebox[1.75em][r]{\deltaabs#1\relax}%
  \else\makebox[0.6em][c]{\ensuremath{\pm}}\makebox[1.75em][r]{0.00}\fi\fi)}
\newcommand{\scorepair}[2]{\begingroup
  \settowidth{\scorewidth}{{\normalfont\bfseries 0.00}}%
  \settowidth{\deltawidth}{{\normalfont\deltasize\printdelta{+0.00}}}%
  \mbox{\makebox[\scorewidth][r]{#1}\,%
    {\normalfont\deltasize\makebox[\deltawidth][l]{#2}}}%
  \endgroup}

\newcommand{\scoredelta}[2]{\scorepair{#1}{%
  \ifdim #2pt>0pt\textcolor{i2vtgain}{\printdelta{#2}}%
  \else\ifdim #2pt<0pt\textcolor{i2vtdrop}{\printdelta{#2}}%
  \else\printdelta{#2}\fi\fi}}
\colorlet{PaperShade}{gray!15}

\iflatexml\else
\AtBeginEnvironment{tabular*}{%
  }
\fi
\newsavebox{\PaperClipBox}
\newenvironment{PaperClipWidth}
  {\begin{lrbox}{\PaperClipBox}}
  {\end{lrbox}\clipbox{0pt 0pt 0pt 0pt}{\usebox{\PaperClipBox}}}
\iflatexml\else
\BeforeBeginEnvironment{tabular*}{\begin{PaperClipWidth}}
\AfterEndEnvironment{tabular*}{\end{PaperClipWidth}}
\fi

\iflatexml
\newcommand{\tabheadrule}{\midrule}
\else
\makeatletter
\newcommand{\tabheadrule}{\noalign{\ifnum0=`}\fi
  \@aboverulesep=\aboverulesep \global\@belowrulesep=\belowrulesep
  \global\@thisruleclass=\@ne
  \@ifnextchar[{\@BTrule}{\@BTrule[1.2pt]}}
\makeatother
\fi

\AtBeginDocument{\captionsetup[table]{position=below,skip=7pt,labelfont=bf}}

\iflatexml
  \renewcommand{\printdelta}[1]{(#1)}
  \renewcommand{\scorepair}[2]{#1\,{\scriptsize #2}}
  
  \providecommand{\Needspace}[1]{}
  \providecommand{\maketitlesupplementary}{\section*{Appendix}}
\fi
\ifnum\PaperBuildSupplement=1\relax\usepackage{xr}\fi
\usepackage[pagebackref,breaklinks,colorlinks,allcolors=cvprblue]{hyperref}
\usepackage[capitalize,noabbrev]{cleveref}
\ifnum\PaperBuildSupplement=1\relax
\fi

\theoremstyle{plain}

\theoremstyle{definition}

\theoremstyle{remark}

\newcommand{\methodname}{I2VT\xspace}

\newcommand{\benchname}{OpenVE++\xspace}

\definecolor{findingaccent}{HTML}{617C96}
\iflatexml\else
\newtcolorbox[auto counter]{findingbox}{
    enhanced,
    sharp corners,
    colback=findingaccent!6,
    colframe=findingaccent!45,
    boxrule=0.5pt,
    left=6pt, right=6pt, top=4pt, bottom=4pt,
    fonttitle=\bfseries\small,
    before upper={\textsc{\small Finding~\thetcbcounter.}\enspace},
}
\fi

\makeatletter
\renewcommand\paragraph{\@startsection{paragraph}{4}{\z@}%
  {6pt \@plus 2pt \@minus 2pt}{-1em}{\normalfont\normalsize\bfseries}}
\makeatother

\ifdefined\PaperShowComments
\newcommand{\lp}[1]{\textcolor{magenta}{LP: #1}}
\newcommand{\xj}[1]{\textcolor{orange}{Ji: #1}}
\newcommand{\yxp}[1]{\textcolor{red}{Xiangpeng: #1}}
\newcommand{\yh}[1]{\textcolor{blue}{Yanhao: #1}}
\else
\newcommand{\lp}[1]{}
\newcommand{\xj}[1]{}
\newcommand{\yxp}[1]{}
\newcommand{\yh}[1]{}
\fi
\def\confName{CVPR}
\def\confYear{2027}

\title{Improving and Expanding Video Editing Capabilities Without Videos}

\ifnum\PaperBuildSupplement=2\relax
  \iflatexml
  \usepackage{authblk}
  \author[1]{Ji Xie}
  \author[2]{Yanhao Jia}
  \author[3]{Xiangpeng Yang}
  \author[1]{Martin Q. Ma}
  \author[4]{Sihan Xu}
  \author[2]{Erik Cambria}
  \author[1]{Louis-Philippe Morency}
  \affil[1]{Carnegie Mellon University}
  \affil[2]{Nanyang Technological University}
  \affil[3]{University of Technology Sydney}
  \affil[4]{University of Michigan}
  \else
  \author{%
    Ji Xie$^{1}$ \quad Yanhao Jia$^{2}$ \quad Xiangpeng Yang$^{3}$ \quad Martin Q. Ma$^{1}$\\[2pt]
    Sihan Xu$^{4}$ \quad Erik Cambria$^{2}$ \quad Louis-Philippe Morency$^{1}$\\[2pt]
    {\normalfont $^{1}$\,Carnegie Mellon University \quad $^{2}$\,Nanyang Technological University}\\[2pt]
    {\normalfont $^{3}$\,University of Technology Sydney \quad $^{4}$\,University of Michigan}\\[2pt]{\normalfont\href{https://i2v-transfer.github.io/}{\texttt{https://i2v-transfer.github.io}}}}
  \fi
  \hypersetup{pdfauthor={Ji Xie, Yanhao Jia, Xiangpeng Yang, Martin Q. Ma, Sihan Xu, Erik Cambria, Louis-Philippe Morency}}
\else
  \author{Anonymous Authors}
\fi

\ifdefined\PaperRevisionLoaded
\else
\def\PaperRevisionLoaded{1}
\providecommand{\PaperShowRevisions}{0}
\usepackage{xcolor}
\usepackage[normalem]{ulem}
\definecolor{PaperRevisionAdded}{HTML}{007C78}
\definecolor{PaperRevisionDeleted}{HTML}{B13A45}
\DeclareRobustCommand{\DIFadd}[1]{%
  \ifnum\PaperShowRevisions=1\relax{\color{PaperRevisionAdded}\relax#1}\else#1\fi}
\let\PaperLiveRef\ref
\let\PaperLiveCref\Cref
\DeclareRobustCommand{\PaperDeletedRef}[1]{%
  \ifcsname r@#1\endcsname\PaperLiveRef{#1}\else[removed]\fi}
\DeclareRobustCommand{\PaperDeletedCref}[1]{%
  \ifcsname r@#1\endcsname\PaperLiveCref{#1}\else[removed reference]\fi}
\DeclareRobustCommand{\DIFdel}[1]{%
  \ifnum\PaperShowRevisions=1\relax{\color{PaperRevisionDeleted}\let\ref\PaperDeletedRef\let\Cref\PaperDeletedCref\hspace{0.12em}\sout{#1}\hspace{0.12em}}\fi}

\providecommand{\DIFdelFL}[1]{\DIFdel{#1}}

\fi

\newcommand{\PaperSupp}{\ifnum\PaperBuildSupplement=2\relax the appendix\else the supplementary material\fi}

\begin{document}
\raggedbottom
\ifnum\PaperBuildSupplement=1\relax
  \clearpage
  \appendix
  \setcounter{table}{0}%
  \setcounter{figure}{0}%
  \setcounter{equation}{0}%
  \setcounter{algorithm}{0}%
  \renewcommand{\thetable}{S\arabic{table}}%
  \renewcommand{\thefigure}{S\arabic{figure}}%
  \renewcommand{\theequation}{S\arabic{equation}}%
  \renewcommand{\thealgorithm}{S\arabic{algorithm}}%
  \def\theHtable{S.\arabic{table}}%
  \def\theHfigure{S.\arabic{figure}}%
  \def\theHequation{S.\arabic{equation}}%
  \def\theHalgorithm{S.\arabic{algorithm}}%
  \maketitlesupplementary
  \begingroup
\ifdefined\nolinenumbers\nolinenumbers\fi
\noindent\begin{minipage}[t]{\columnwidth}
\normalfont\normalsize\raggedright
\hypersetup{linkcolor=black}
\newcommand{\appendixentry}[2]{%
  \noindent\hyperref[#1]{\makebox[\linewidth][l]{%
    \makebox[1.5em][l]{\textbf{\ref*{#1}.}}%
    #2\hspace{0.5em}%
    \leaders\hbox to .55em{\hss\textcolor{black!28}{.}\hss}\hfill
    \hspace{0.5em}\textcolor{black!60}{\pageref*{#1}}}}\par}
\noindent\textbf{Appendix Contents}\par
\vspace{4pt}
  \small\setlength{\parskip}{2pt}
  \appendixentry{app:discussion}{Discussion}
  \appendixentry{app:impl}{Implementation Details}
  \appendixentry{app:data}{Training Data Construction}
  \appendixentry{app:full_scores}{Additional OpenVE-Bench Results}
  \appendixentry{app:ivebench}{Generalization to IVEBench}
  \appendixentry{app:ablation}{More Ablations}
  \appendixentry{app:extended}{\benchname}
  \appendixentry{app:user_study}{User Study}
  \appendixentry{app:evaluation_protocol}{Evaluation Details}
  \appendixentry{app:eval_prompts}{Evaluation Prompts}
  \appendixentry{app:videocof_gallery}{Additional VideoCoF Results}
\end{minipage}\par
\endgroup
\vspace{0.7em}

  \PaperInput{sections/A_appendix}
\else
\twocolumn[{
\renewcommand\twocolumn[1][]{#1}
\maketitle
\iflatexml\begin{center}\href{https://i2v-transfer.github.io/}{https://i2v-transfer.github.io}\end{center}\fi
\begin{center}
\includegraphics[width=\textwidth]{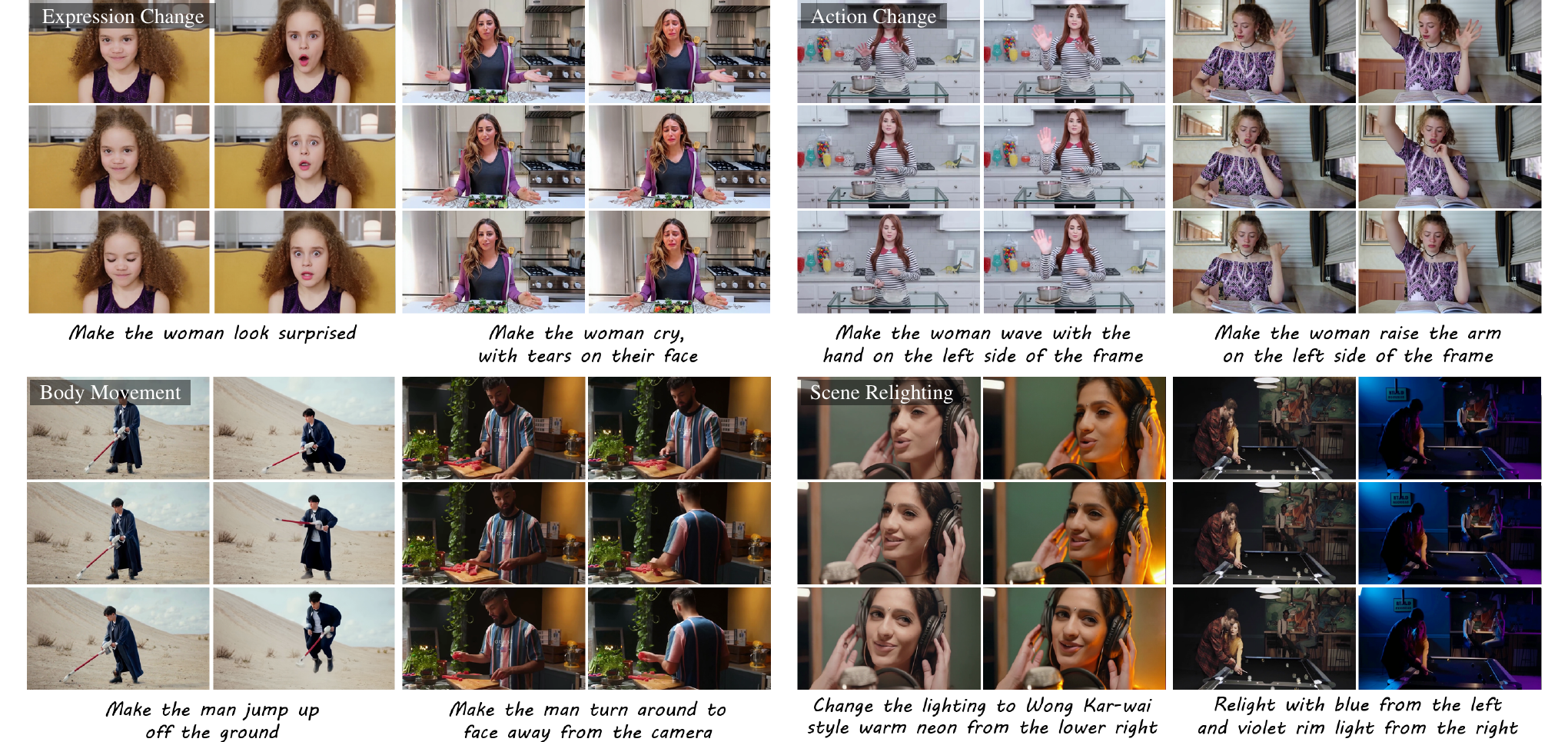}
\captionsetup{hypcap=false}
\captionof{figure}{\textbf{Video editing capabilities learned from images.} \methodname expands video editors to new editing tasks, including action changes, expression changes, body movements, and scene relighting. Each pair shows frames from the source video on the left and the video edited by our \methodname model on the right, with the editing instruction below.
}
\label{fig:teaser}

\end{center}
\vspace{0.5em}
}]

\begin{abstract}
\PaperInput{sections/0_abstract}
\end{abstract}

\PaperInput{sections/1_introduction}
\PaperInput{sections/3_method}
\PaperInput{sections/4_experiments}
\PaperInput{sections/2_related_work}
\PaperInput{sections/5_conclusion}

\FloatBarrier
\fi

\ifnum\PaperBuildSupplement=0\relax
\section*{LLM Usage Statement}
We used Claude Code for coding assistance and Claude and Gemini to polish the paper. In evaluation, the VLM Gemini-2.5-Pro rates every edited video on both benchmarks (\Cref{sec:evaluation_setting}). In training-data construction, VLMs, including Qwen3.5-27B, inspect source frames and write and rewrite edit instructions.
\fi

{
\small
\bibliographystyle{styles/ieeenat_fullname_sources}
\bibliography{references}
}

\ifnum\PaperBuildSupplement=2\relax
  \clearpage
  \appendix
  \setcounter{table}{0}%
  \setcounter{figure}{0}%
  \setcounter{equation}{0}%
  \setcounter{algorithm}{0}%
  \renewcommand{\thetable}{S\arabic{table}}%
  \renewcommand{\thefigure}{S\arabic{figure}}%
  \renewcommand{\theequation}{S\arabic{equation}}%
  \renewcommand{\thealgorithm}{S\arabic{algorithm}}%
  \def\theHtable{S.\arabic{table}}%
  \def\theHfigure{S.\arabic{figure}}%
  \def\theHequation{S.\arabic{equation}}%
  \def\theHalgorithm{S.\arabic{algorithm}}%
  \maketitlesupplementary
  \begingroup
\ifdefined\nolinenumbers\nolinenumbers\fi
\noindent\begin{minipage}[t]{\columnwidth}
\normalfont\normalsize\raggedright
\hypersetup{linkcolor=black}
\newcommand{\appendixentry}[2]{%
  \noindent\hyperref[#1]{\makebox[\linewidth][l]{%
    \makebox[1.5em][l]{\textbf{\ref*{#1}.}}%
    #2\hspace{0.5em}%
    \leaders\hbox to .55em{\hss\textcolor{black!28}{.}\hss}\hfill
    \hspace{0.5em}\textcolor{black!60}{\pageref*{#1}}}}\par}
\noindent\textbf{Appendix Contents}\par
\vspace{4pt}
  \small\setlength{\parskip}{2pt}
  \appendixentry{app:discussion}{Discussion}
  \appendixentry{app:impl}{Implementation Details}
  \appendixentry{app:data}{Training Data Construction}
  \appendixentry{app:full_scores}{Additional OpenVE-Bench Results}
  \appendixentry{app:ivebench}{Generalization to IVEBench}
  \appendixentry{app:ablation}{More Ablations}
  \appendixentry{app:extended}{\benchname}
  \appendixentry{app:user_study}{User Study}
  \appendixentry{app:evaluation_protocol}{Evaluation Details}
  \appendixentry{app:eval_prompts}{Evaluation Prompts}
  \appendixentry{app:videocof_gallery}{Additional VideoCoF Results}
\end{minipage}\par
\endgroup
\vspace{0.7em}

  \PaperInput{sections/A_appendix}
\fi

\end{document}